\documentclass{article}
\usepackage{ijcai26}

\usepackage{times}
\usepackage{soul}
\usepackage{url}
\usepackage[hidelinks]{hyperref}
\usepackage[utf8]{inputenc}
\usepackage[small]{caption}
\usepackage{subfig}
\usepackage{graphicx}
\usepackage{amsmath}
\usepackage{amsthm}
\usepackage{amsfonts} 
\usepackage{booktabs}
\usepackage{algorithm}
\usepackage{algorithmic}
\usepackage[switch]{lineno}
\usepackage{lipsum}

\usepackage{xcolor}

\title{\textsc{FloorplanVLM}: A Vision-Language Model for Floorplan Vectorization}

\author{
Yuanqing Liu\thanks{Equal contribution.}
\and
Ziming Yang\footnotemark[1]\and
Yulong Li\and
Yue Yang\thanks{Corresponding Author.}
\affiliations
Beike\\
\emails
\{liuyuanqing006, yangziming006, liyulong008, yangyue092\}@ke.com,
}

\begin{document}

\maketitle
\begin{abstract}
    Converting raster floorplans into engineering-grade vector graphics is challenging due to complex topology and strict geometric constraints. To address this, we present \textsc{FloorplanVLM}, a unified framework that reformulates floorplan vectorization as an image-conditioned sequence modeling task. Unlike pixel-based methods that rely on fragile heuristics or query-based transformers that generate fragmented rooms, our model directly outputs structured JSON sequences representing the global topology. This ``pixels-to-sequence'' paradigm enables the precise and holistic constraint satisfaction of complex geometries, such as slanted walls and curved arcs. To support this data-hungry approach, we introduce a scalable data engine: we construct a large-scale dataset (\textsc{Floorplan-2M}) and a high-fidelity subset (\textsc{Floorplan-HQ-300K}) to balance geometric diversity and pixel-level precision. We then employ a progressive training strategy, using Supervised Fine-Tuning (SFT) for structural grounding and quality annealing, followed by Group Relative Policy Optimization (GRPO) for strict geometric alignment. To standardize evaluation on complex layouts, we establish and open-source \textsc{FPBench-2K}. Evaluated on this rigorous benchmark, \textsc{FloorplanVLM} demonstrates exceptional structural validity, achieving \textbf{92.52\%} external-wall IoU and robust generalization across non-Manhattan architectures.
\end{abstract}

\begin{figure}
    \centering
    \includegraphics[width=\linewidth]{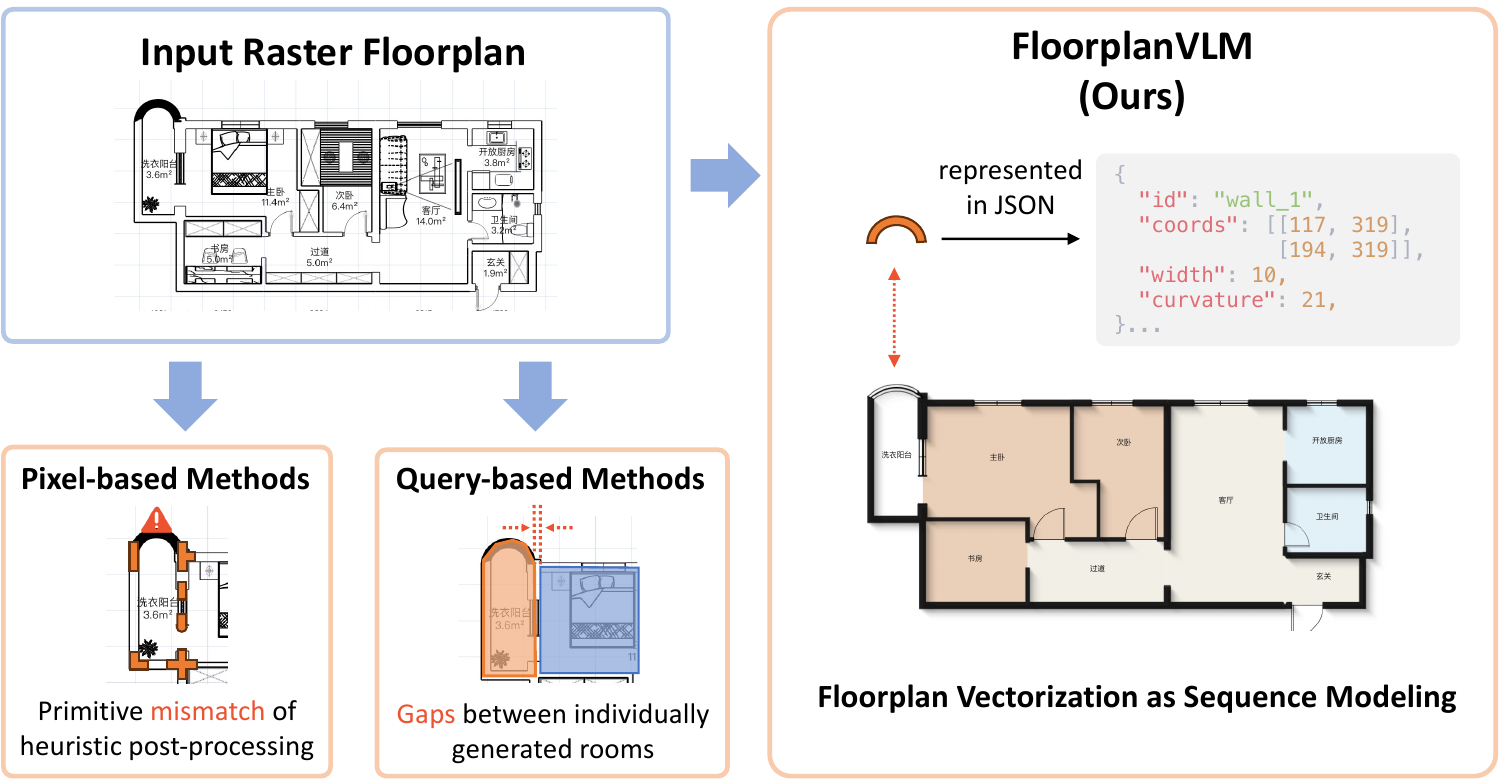}
    \caption{\textbf{Comparison of Floorplan Vectorization Paradigms.} Unlike pixel-based methods that lead to primitive mismatches or query-based methods that cause topological gaps, \textsc{FloorplanVLM} reframes vectorization as sequence modeling. The model directly outputs structured JSON sequences containing geometric attributes like coordinates and curvature, enabling high-quality floorplan reconstruction.}
    \label{fig:comparison}
\end{figure}

\section{Introduction}

Floorplan vectorization requires transforming rasterized images into structured geometric primitives (e.g., walls and rooms) while maintaining strict topological consistency~\cite{gupta2024advancements}. Unlike natural image captioning, this task demands precise geometric alignment and logical correctness—walls must connect, and rooms must form closed loops. Traditional workflows heavily rely on manual tracing, creating a labor-intensive bottleneck for scalable architectural modeling~\cite{skrzypczak2022scan,salah2025architectural}.

Existing approaches typically treat vectorization as segmentation or isolated polygon detection~\cite{kalervo2019cubicasa5k,wang2022roomformer}. While effective on simple layouts, these paradigms face significant challenges in scalability and structural coherence. As illustrated in Figure~\ref{fig:comparison}, pixel-based methods often struggle with resolution limits and require complex heuristics to assemble vector graphs. Similarly, query-based methods generating independent room polygons often suffer from structural decoupling to resolve shared boundaries in complex, non-Manhattan designs.

In this work, we introduce \textsc{FloorplanVLM}, a novel framework that reformulates vectorization as an \textit{image-conditioned sequence modeling task}. Diverging from pixel-based methods that rely on fragile heuristics and query-based approaches that suffer from structural decoupling, our model directly synthesizes a unified JSON sequence describing the global topology. Central to this paradigm is a dependency-ordered serialization strategy: by explicitly defining a shared wall skeleton first and subsequently characterizing rooms as references to these structural primitives, we enforce topological consistency by design. This strategy eliminates the geometric gaps inherent in independent polygon predictions, while effectively capturing complex non-Manhattan geometries, such as slanted walls and curved arcs.

However, a fundamental tension exists: the probabilistic token generation of VLMs inherently conflicts with the deterministic precision required for architectural engineering. We resolve this by integrating a robust Data Engine with a geometric alignment strategy. Addressing the scarcity of diverse training data, we introduce a scalable Data Engine. We observe that raw industrial data, while abundant, often suffers from \textit{coordinate misalignment} between raster screenshots and real-world structural labels. To leverage this scale, we first construct \textsc{Floorplan-2M} via structure-aware clustering to capture long-tail geometric distributions despite spatial noise. We then bridge the precision gap by distilling \textsc{Floorplan-HQ-300K}, a hybrid of human-recaptioned and synthetic re-rendered samples, to ensure strict \textit{pixel-aligned} grounding.

Our contributions are summarized as follows:
\begin{itemize}
    \item \textbf{End-to-End Sequence Modeling:} We propose a framework that reformulates vectorization as image-conditioned sequence generation, directly outputting render-ready JSON floorplans and enabling the holistic reconstruction of complex layouts without heuristic post-processing.
    \item \textbf{Scalable Data Engine \& Open Benchmark:} We introduce a hierarchical data strategy, leveraging the large-scale \textsc{Floorplan-2M} for structural diversity and the re-rendered and re-labeled \textsc{Floorplan-HQ-300K} for \textit{pixel-aligned} precision. Furthermore, we establish and release \textsc{FPBench-2K}, a rigorous \textbf{open-source} benchmark encompassing various complex floorplans, providing the community with a standardized testbed for topological reasoning.
    \item \textbf{Geometric Alignment via RL:} We bridge the non-differentiable gap between discrete tokens and continuous geometry using GRPO, reducing spatial hallucinations and achieving \textbf{92.52\%} External Wall IoU on \textsc{FPBench-2K}.
\end{itemize}

\section{Related Work}

\subsection{Floorplan Vectorization}
Floorplan vectorization has evolved from traditional low-level image processing to two dominant deep-learning paradigms: pixel-based and query-based methods.

\paragraph{Pixel-based Approaches.} 
Methods such as CubiCasa5K~\cite{kalervo2019cubicasa5k} and L-CNN~\cite{zhou2019end} treat vectorization as segmentation or heatmap detection. While effective at extracting local features, they exhibit a fundamental \textit{representation mismatch}: they output discrete raster maps while vectorization requires continuous graphs. Consequently, these methods rely on heavy, heuristic post-processing rules to assemble vector primitives, which often fail when encountering non-Manhattan geometries (e.g., slanted walls) that defy rigid connectivity assumptions.

\paragraph{Query-based Approaches.} 
More recent works (e.g., RoomFormer~\cite{wang2022roomformer} and PolyRoom~\cite{liu2024polyroom}) utilize Transformers with fixed-size queries to predict vector sets. However, these methods typically operate in a \textit{room-wise manner}, generating individual polygons independently. While these models capture global context, their room-centric parameterization fails to enforce the uniqueness of shared structural elements. This frequently results in structural inconsistencies—such as gaps between shared walls or overlapping rooms—that require complex, error-prone merging algorithms to resolve.

\subsection{Vision-Language Models for Geometric Generation}
Recent Large Vision-Language Models (VLMs)~\cite{openai2023gpt4,bai2025qwen2} have demonstrated impressive semantic reasoning but often struggle with precise spatial generation, a phenomenon known as ``geometric hallucination''~\cite{chen2025survey}. While adaptation efforts like ChartLlama~\cite{han2023chartllama} have applied VLMs to structured data extraction (e.g., charts), architectural drawings impose significantly stricter constraints: outputs must not only be syntactically valid JSON but also geometrically watertight (e.g., closed loops, aligned coordinates). Current literature lacks a scalable data synthesis framework capable of grounding the probabilistic generation of VLMs into the strict, verifiable geometric constraints required for engineering-grade vectorized floorplan reconstruction.

\subsection{Reinforcement Learning for Geometric Alignment}
While Reinforcement Learning from Human Feedback (RLHF)~\cite{ouyang2022training} excels at aligning models with subjective human preferences, geometric vectorization demands adherence to strict, objective constraints. This requirement aligns with recent advancements in mathematical reasoning, where frameworks like DeepSeekMath~\cite{shao2024deepseekmath} introduce Reinforcement Learning with Verifiable Rewards (RLVR) to optimize for deterministic correctness (e.g., answer compilation) rather than learned reward proxies. Extending this paradigm to the geometric domain, recent works such as SpaceR~\cite{ouyang2025spacer} and ReCAD~\cite{li2025recad} utilize geometric fidelity metrics as reward signals. These approaches demonstrate the capability of RLVR to transform VLMs from probabilistic token generators into rigorous geometric reasoners.

\begin{figure*}[t]
    \centering
    \includegraphics[width=\linewidth]{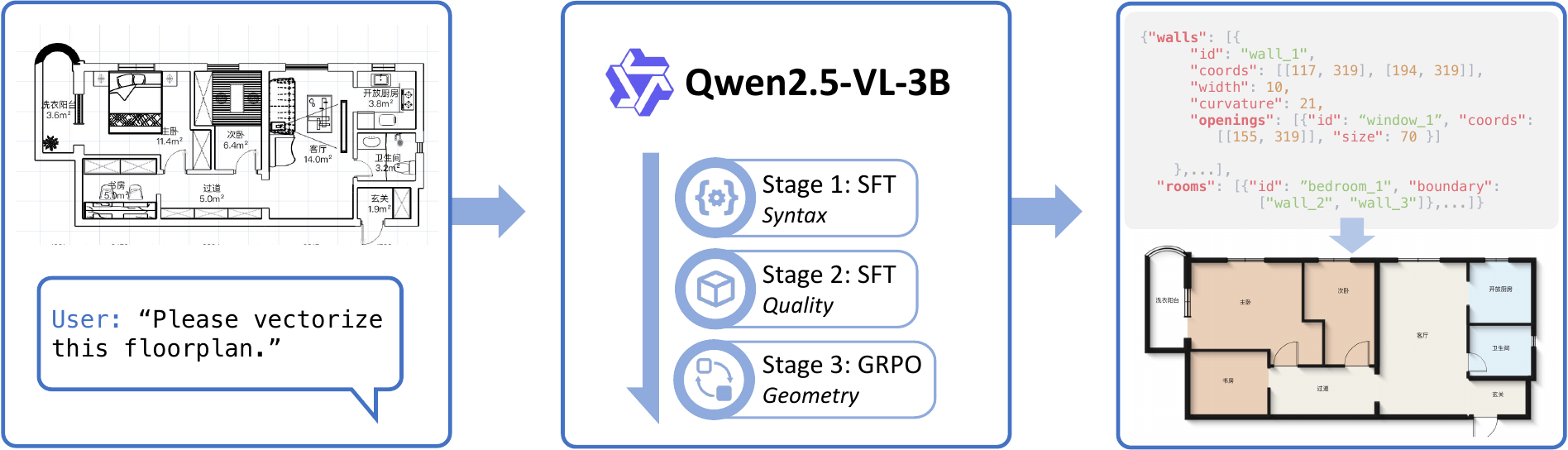}
    \caption{
        \textbf{Overview of \textsc{FloorplanVLM} Framework.} 
        We reformulate floorplan vectorization as a multi-modal sequence generation task using Qwen2.5-VL. 
        \textbf{(Left)} The model accepts a raster floorplan and a prompt, transforming visual data into a discrete token sequence.
        \textbf{(Center)} A progressive training pipeline bridges the gap between visual perception and geometric logic: Stages 1 \& 2 (SFT) establish syntactic correctness and generation quality, while Stage 3 employs Group Relative Policy Optimization (GRPO) to enforce strict geometric alignment (e.g., closed loops).
        \textbf{(Right)} The output is a structured, hierarchical JSON sequence that deterministically parses into a precise vector floorplan.
    }
    \label{fig:overview}
\end{figure*}

\section{Problem Formulation}\label{sec:problem_formulation}

\subsection{Sequence Modeling}

We cast floorplan vectorization as an image-conditioned sequence modeling task. Given a rasterized floorplan image $I$, our goal is to generate a sequence of discrete tokens $T = (t_1, t_2, \dots, t_L)$ that deterministically parses into a structured geometric graph $\mathcal{S}$. We model the conditional probability $P_\theta(T|I)$ autoregressively:

\begin{equation}
P_\theta(T|I) = \prod_{j=1}^{L} P(t_j \mid t_{j-1}, \dots, t_1, I),
\end{equation}
where $\theta$ denotes the learnable parameters of the model. The token sequence $T$ is deterministically converted into the structured geometric graph $\mathcal{S}$ via a parsing function, i.e., $\mathcal{S} = \texttt{parse}(T)$.

\subsection{Geometric Alignment}
While sequence modeling maximizes token likelihood, our ultimate goal is to ensure the reconstructed geometry $\mathcal{S}$ aligns with the ground truth structure $\mathcal{S}_{gt}$. We formalize this as minimizing a geometric discrepancy metric $D$:

\begin{equation} \label{geo_obj}
\theta^* = \arg\min_{\theta} \mathbb{E}_{T \sim P_\theta(\cdot|I)} [D(\texttt{parse}(T), \mathcal{S}_{gt})],
\end{equation}
where $D$ is a composite metric that evaluates both geometric fidelity (e.g., Intersection over Union) and topological validity. Crucially, the transformation $\texttt{parse}(\cdot)$ and the metric $D(\cdot)$ involve non-differentiable operations. This creates a fundamental disconnect between token-level training objectives and geometry-level evaluation, necessitating an optimization strategy beyond standard maximum likelihood estimation.

\subsection{Floorplan Representation}
To ensure topological consistency and support complex non-Manhattan geometries, we define $\mathcal{S} = (\mathcal{W}, \mathcal{O}, \mathcal{R})$ via a dependency-ordered parameterization.

\paragraph{Walls.} The floorplan skeleton is a set of walls $\mathcal{W} = \{w_1, \dots, w_N\}$. We parameterize each wall $w_i$ to accommodate curved and slanted geometries:
\begin{equation}
w_i = (\mathbf{p}_{start}, \mathbf{p}_{end}, \tau, \kappa, \mathcal{O}_i),
\end{equation}
where $\mathbf{p}_{start}, \mathbf{p}_{end} \in \mathbb{R}^2$ denote the endpoint coordinates, and $\tau \in \mathbb{R}^+$ represents the wall thickness. The curvature parameter $\kappa$ defines the wall geometry: $\kappa = 0$ indicates a straight wall, while $\kappa \neq 0$ indicates a curved arc connecting the endpoints; the sign determines the curvature direction relative to the wall vector.

\paragraph{Openings.} Openings (doors and windows) are modeled as attributes nested within their parent walls rather than as independent entities. The set of openings $\mathcal{O}_i$ associated with wall $w_i$ contains elements parameterized as:
\begin{equation}
o = (c, \delta, \omega),
\end{equation}
where $c \in \{\texttt{door}, \texttt{window}\}$ denotes the semantic class, $\omega$ is the opening width, and $\delta$ specifies the coordinate of the opening center along the wall centerline. This hierarchical design constrains openings to lie on their parent walls, significantly reducing geometric inconsistencies.

\paragraph{Rooms.} We define rooms as functional zones derived from the wall graph. Each room $r_k \in \mathcal{R}$ is characterized by:
\begin{equation}
r_k = (\ell, \mathcal{E}_k),
\end{equation}
where $\ell$ is the semantic label (e.g., \texttt{bedroom}), and $\mathcal{E}_k = (i_1, i_2, \dots, i_m)$ is an ordered sequence of indices that reference walls in $\mathcal{W}$ and form a closed topological cycle.

\begin{figure*}[t]
    \centering
    \includegraphics[width=\textwidth]{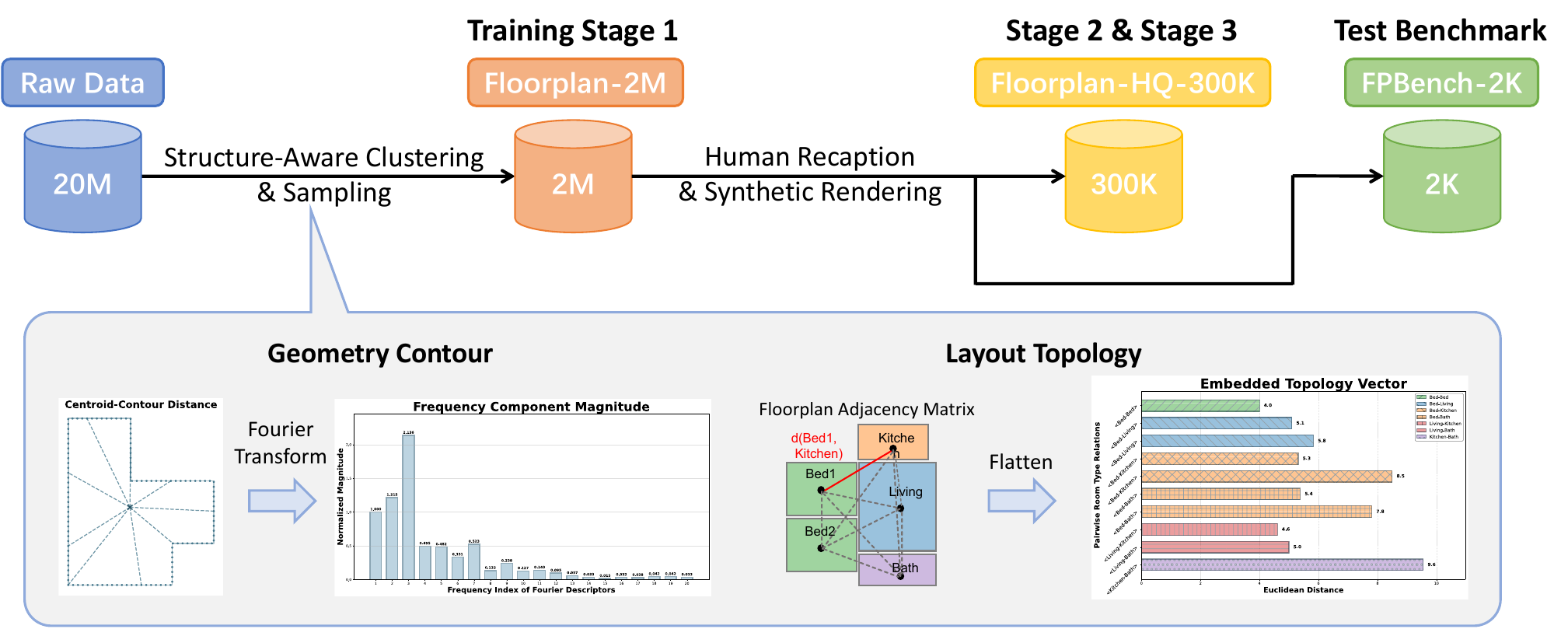}
    \caption{\textbf{Overview of the Scalable Data Engine.} 
The pipeline illustrates the construction of our hierarchical dataset structure.
\textbf{(Top)} Starting from a raw pool of 20M samples, we construct \textsc{Floorplan-2M} via structure-aware clustering to ensure geometric diversity. Subsequently, we curate the high-fidelity \textsc{Floorplan-HQ-300K} subset through a hybrid process combining human recaption and synthetic rendering (for pixel-perfect alignment), serving as the foundation for Stage 2 \& 3 training.
\textbf{(Bottom)} To enable balanced sampling, we extract dual-view features: \textbf{Geometry Contour} utilizes Fourier Descriptors to capture global boundary invariance, while \textbf{Layout Topology} employs graph embeddings to encode the internal spatial logic of room connections.}
    \label{fig:data_engine}
\end{figure*}

\section{Methodology}

\subsection{Framework Overview}
\label{sec:framework_overview}

\textsc{FloorplanVLM} is a unified generative framework designed to transcribe rasterized architectural drawings into precise, topology-aware vector code. To address the challenge of aligning probabilistic token generation with strict geometric constraints (as formulated in Sec.~\ref{sec:problem_formulation}), we integrate three core components:

\begin{enumerate}
\item \textbf{Sequence Serialization (Sec.~\ref{sec:serialization}):} We introduce a token-efficient JSON serialization schema that maps hierarchical geometric primitives (walls, rooms, and openings) into a discrete sequence, using a custom vocabulary to reduce sequence length and improve stability.
\item \textbf{Data Engine (Sec.~\ref{sec:data_engine}):} To support data-hungry VLM training, we introduce a scalable pipeline that constructs the \textsc{Floorplan-2M} dataset via topological clustering and curates a high-fidelity subset, \textsc{Floorplan-HQ-300K}, for precision refinement.
\item \textbf{Progressive Training (Sec.~\ref{sec:training}):} We employ a three-stage optimization strategy. The model first learns syntactic rules and visual patterns via SFT, and then transitions to GRPO, a reinforcement learning algorithm that directly optimizes the non-differentiable geometric objectives defined in Eq.~\eqref{geo_obj}.
\end{enumerate}

Figure~\ref{fig:overview} illustrates the overall pipeline, with Qwen2.5-VL-3B as the foundational vision-language model~\cite{bai2025qwen2}.

\subsection{Geometric Sequence Serialization}
\label{sec:serialization}

To enable the VLM to generate precise geometric vector graphics, we bridge the modality gap between continuous 2D coordinates and discrete 1D text tokens. We propose a token-efficient serialization strategy that maps the hierarchical layout $\mathcal{S}$ (defined in Sec.~\ref{sec:problem_formulation}) into a compact JSON sequence.

\subsubsection{JSON Schema}
To strictly enforce topological consistency, we adopt a declarative JSON schema with a \textit{``Structure-First, Semantics-Second''} serialization order, as shown in Figure~\ref{fig:overview}:

\begin{enumerate}
    \item \textbf{Geometric Skeleton ($\mathcal{W}$):} The sequence begins by defining all walls and their nested openings. Each wall $w_i$ is assigned a unique identifier (e.g., \texttt{"wall\_1"}) and explicit attributes, including coordinates, curvature $\kappa$, thickness $\tau$, and its openings $\mathcal{O}_i$.
    \item \textbf{Functional Zones ($\mathcal{R}$):} Rooms are then defined by referencing the identifiers of their enclosing walls (e.g., \texttt{["wall\_1", "wall\_2"]}).
\end{enumerate}

This dependency-ordered approach prevents common topological errors, such as floating rooms or gaps between shared boundaries, because each room definition is strictly grounded in the pre-defined wall skeleton.

\subsubsection{Token Compression and Representation}
Standard JSON is verbose, which can destabilize long-sequence generation. We address this via a hybrid representation strategy:

\paragraph{Coordinate Normalization.} To align geometric data with the visual encoder resolution, we normalize all floorplan coordinates by scaling the longer image edge to $1024$ while preserving the aspect ratio. These coordinates are serialized as plain text tokens. This approach leverages the VLM's inherent capability to process numerical text, avoiding the complexity of defining a separate positional vocabulary while maintaining sufficient precision for architectural reconstruction.

\paragraph{Semantic Token Compression.} While coordinates remain plain text, we introduce a vocabulary of 1,391 custom special tokens to encode high-frequency JSON syntax and semantic keys. For instance, repeated attribute keys such as \texttt{"curvature"} or \texttt{"openings"} are compressed into single tokens (e.g., \texttt{<cv>}, \texttt{<op>}). This strategy reduces the total sequence length by approximately 25\%, significantly lowering computational cost and ensuring the model operates within a manageable context window.

\begin{figure}[t]
    \centering
    \subfloat[Geometry distribution \\of \textsc{Floorplan-2M}.]{
        \includegraphics[width=0.48\linewidth]{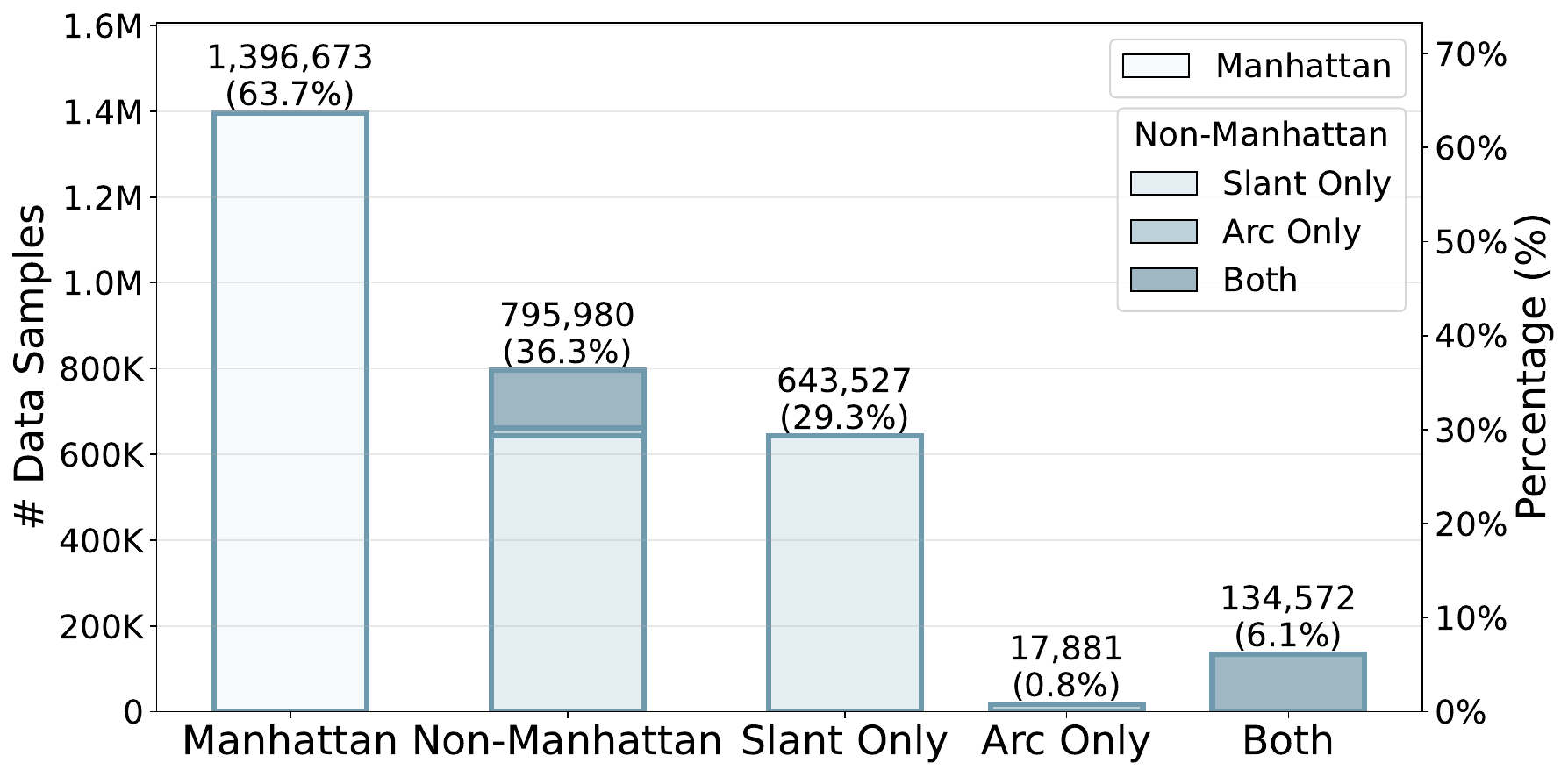}
        \label{fig:data_stats-a}
    }
    \subfloat[Primitive Count Distribution in \textsc{Floorplan-2M}.]{
        \includegraphics[width=0.48\linewidth]{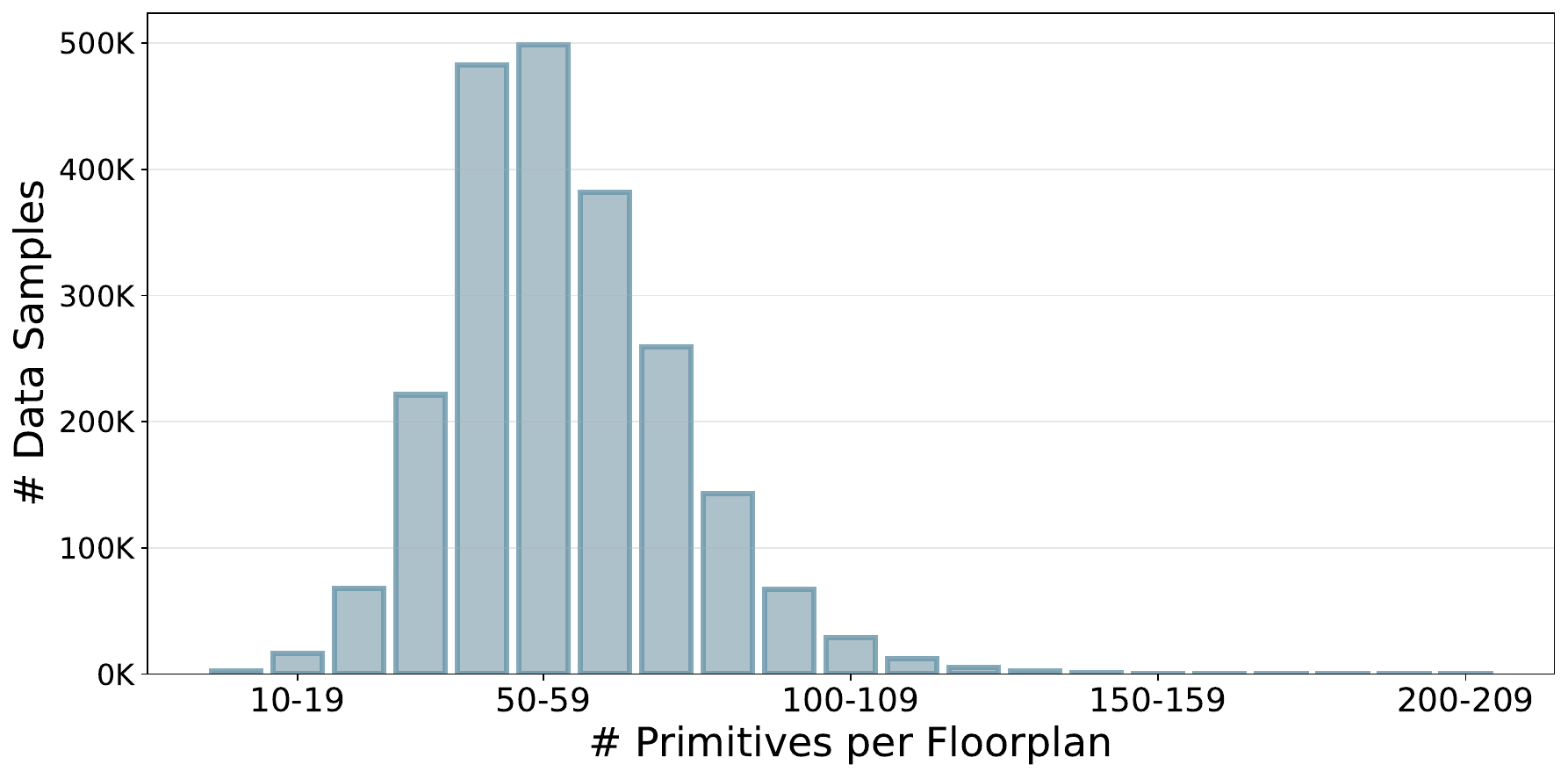}
        \label{fig:data_stats-b}
    }
    
    \subfloat[Geometry distribution \\of \textsc{Floorplan-HQ-300K}.]{
        \includegraphics[width=0.48\linewidth]{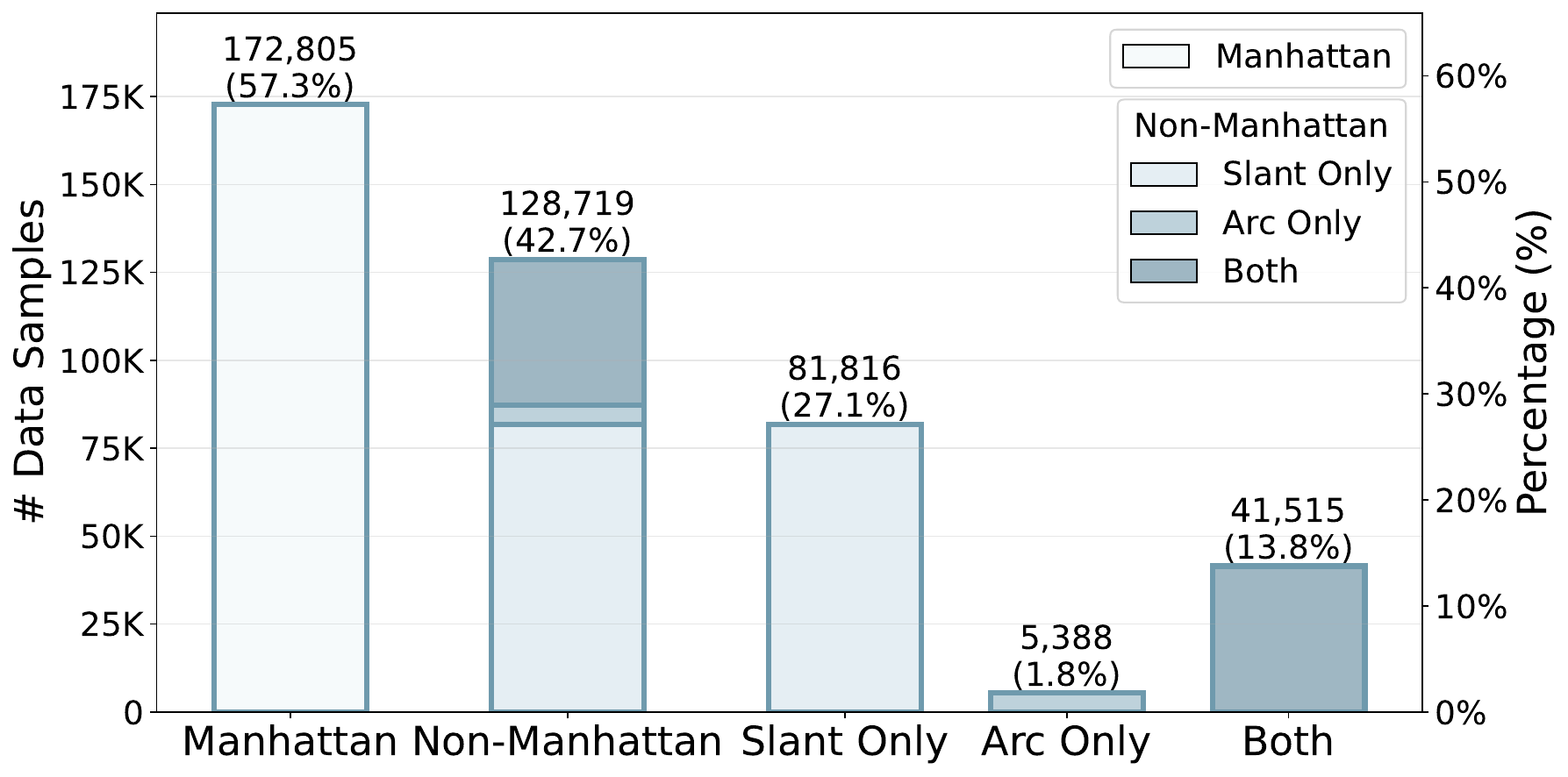}
        \label{fig:data_stats-c}
    }
    \subfloat[Primitive Count Distribution in \textsc{Floorplan-HQ-300K}.]{
        \includegraphics[width=0.48\linewidth]{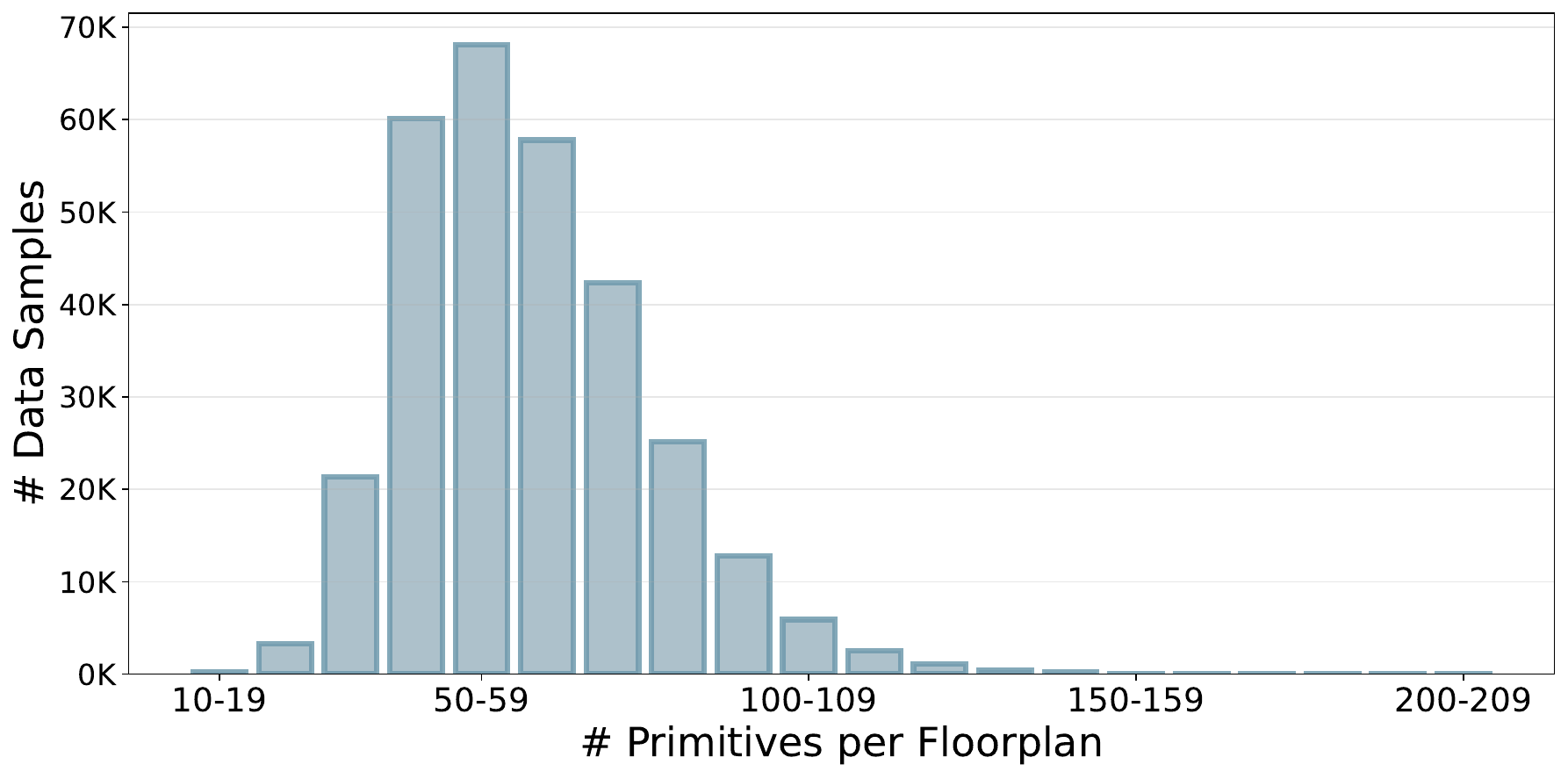}
        \label{fig:data_stats-d}
    }
    \caption{\textbf{Statistical Distribution of Training Datasets.} 
        We compare the geometric type \textbf{(left)} and primitive count \textbf{(right)} distributions between the raw \textsc{Floorplan-2M} \textbf{(top)} and the refined \textsc{Floorplan-HQ-300K} \textbf{(bottom)}. 
        Our structure-aware sampling effectively rebalances the dataset, significantly increasing the proportion of non-Manhattan geometries (from 36.3\% to 42.7\%) while preserving the long-tail distribution of scene complexity.
    }
    \label{fig:data_stats}
\end{figure}

\subsection{Scalable Data Engine}
\label{sec:data_engine}

Training a generalist VLM requires data at scale. However, existing datasets are limited in both quantity and diversity. We introduce a scalable data engine that constructs \textsc{Floorplan-2M}, the largest dataset to date, and refines it into a high-fidelity subset, \textsc{Floorplan-HQ-300K}, for precision tuning.

\subsubsection{\textsc{Floorplan-2M}: Structure-Aware Clustering}
We start with a pool of over 20 million raw vector floorplans from an industrial interior design platform. Naive random sampling would bias the dataset toward repetitive, standard layouts (e.g., Manhattan-style apartments). To better cover long-tail geometries (e.g., slanted walls and curved balconies), we propose an unsupervised \textit{Structure-Aware Clustering} framework:

\paragraph{Feature Extraction.} We characterize floorplan similarity from two complementary perspectives:
\begin{enumerate}
    \item \textbf{Geometric Contour:} We utilize Fourier Contour Descriptors (FCD) \cite{zhang2002comparative} to capture the global shape of the external boundary. By transforming the boundary's 1D shape signature into the frequency domain, we obtain a feature vector invariant to translation, rotation, and scaling.
    \item \textbf{Internal Spatial Structure:} To encode room-arrangement logic, we construct a dual graph where nodes represent functional zones (e.g., bedrooms) and edges represent adjacency. We extract a layout vector based on the sorted pairwise distances between room centroids, making it invariant to room permutation.
\end{enumerate}

\paragraph{Clustering and Sampling.} Using a weighted concatenation of these features, we perform Hierarchical Agglomerative Clustering (HAC)~\cite{mullner2011modern}. We upsample rare clusters and downsample dominant ones, yielding a balanced \textsc{Floorplan-2M} dataset with 2 million pairs. 

It is important to note that the raw images are sourced as \textit{screenshots} from the design platform, while the ground truth labels rely on \textit{real-world structural coordinates} (e.g., in millimeters). Since the spatial transformation (scale and shift) of the screenshots are unknown, a \textit{coordinate gap} exists between the visual pixels and the geometric labels. Despite this alignment noise, this scale enables the model to learn \textit{generalized} structural topology, even if pixel-perfect precision is limited by the source data.

\subsubsection{\textsc{Floorplan-HQ-300K}: High-Precision Refinement}

While \textsc{Floorplan-2M} provides scale, the aforementioned coordinate gap precludes engineering-grade training. To support high-precision alignment, we curate \textsc{Floorplan-HQ-300K}, where all samples possess strictly \textit{pixel-aligned} coordinates:

\begin{itemize}
\item \textbf{Human-Recaptioned Subset (20K):} We sample diverse raw images and employ professional designers to manually redraw the vector graphs over the raster background. This ensures semantic correctness and eliminates the coordinate gap through human visual alignment.
\item \textbf{Synthetic-Rendered Subset (280K):} To scale up pixel-perfect data, we leverage the structural vector graphs from \textsc{Floorplan-2M} but discard the original screenshots. Instead, we re-render these vectors using an internal engine into CAD-style images. Since the rendering process provides access to the exact transformation matrix (scale and shift), we analytically project the real-world structural coordinates into the image pixel space, guaranteeing mathematical watertightness.
\end{itemize}

By combining human-corrected real data with geometrically precise synthetic data, this 300K subset serves as a rigorous ``Gold Standard'' for fine-tuning the model's geometric precision.

\paragraph{Statistical Distribution Analysis.}
To quantitatively verify the impact of our data strategy, we report the statistical distributions of both datasets in Figure~\ref{fig:data_stats}. As shown in Figure~\ref{fig:data_stats}(a), the raw \textsc{Floorplan-2M} is dominated by standard Manhattan layouts (63.7\%). However, our structure-aware active sampling effectively rebalances the training distribution. In the resulting \textsc{Floorplan-HQ-300K} (Figure~\ref{fig:data_stats}(c)), the proportion of \textit{non-Manhattan geometries} (e.g., slanted walls and arcs) increases from 36.3\% to \textbf{42.7\%}. Crucially, the distribution of geometric primitive counts (Figure~\ref{fig:data_stats}(d)) remains consistent with the source long-tail distribution, confirming that our watertight filtering retains structurally complex samples rather than biasing toward simple layouts. This rebalanced, high-complexity training set is essential for the model to generalize beyond simple rectangular forms.

\subsection{Progressive Training Pipeline}
\label{sec:training}

Generating engineering-grade floorplans requires the model to master two distinct capabilities: understanding JSON syntax and reasoning under strict geometric constraints. We propose a three-stage progressive training regimen that transitions from broad semantic understanding to precise geometric alignment.

\subsubsection{Stages 1 \& 2: Supervised Fine-Tuning (SFT)}
The initial phases focus on likelihood maximization to ground the visual encoder and align the language model with our custom serialization format.

\paragraph{Stage 1: Structural Grounding.} We first train the model on the large-scale \textsc{Floorplan-2M} dataset. Due to the coordinate misalignment inherent in the screenshot-based data, explicitly optimizing for pixel-perfect vertex placement is infeasible at this stage. Instead, the primary objective is to adapt the VLM to the broad visual-structural distribution of architectural drawings. 
This stage grounds visual patterns to approximate geometric primitives, establishing a generalized layout understanding before precision refinement.

\paragraph{Stage 2: Quality Annealing.} To reduce hallucinations and stabilize the output distribution, we perform a second round of SFT on the high-fidelity \textsc{Floorplan-HQ-300K} subset.

In both stages, we optimize the standard autoregressive cross-entropy loss:
\begin{equation}
\mathcal{L}_{\text{SFT}}(\theta) = -\sum_{j=1}^{L} \log P_\theta(t_j \mid t_{<j}, I).
\end{equation}

While SFT provides a strong initialization, it optimizes for token-level prediction accuracy, which correlates poorly with holistic geometric validity (e.g., a single wrong coordinate token can break a wall's connectivity).

\subsubsection{Stage 3: Geometric Alignment via GRPO}
To enforce hard physical constraints, we employ reinforcement learning in the final stage. We adopt Group Relative Policy Optimization (GRPO)~\cite{shao2024deepseekmath}, a variant of PPO~\cite{schulman2017proximal} that eliminates the need for a separate value network, thereby reducing memory overhead and improving training stability.

For each input image $I$, we sample a group of $G$ outputs $\{T_k\}_{k=1}^{G}$ from the old policy $\pi_{\theta_\text{old}}$. Let $\rho(T)=\frac{\pi_{\theta}(T\mid I)}{\pi_{\theta_\text{old}}(T\mid I)}$ be the policy ratio. We optimize the current model $\pi_\theta$ by maximizing the advantage of outputs with higher geometric rewards:

\begin{align}
    &\mathcal{J}_{\text{GRPO}}(\theta) = \mathbb{E}_{I\sim\mathcal{D},\{T_k\}_{k=1}^G \sim \pi_{\theta_\text{old}}(\cdot\mid I)} \notag\\
    &\!\!\! \left[\frac{1}{G} \sum_{k=1}^{G} \big(\min \left( A_k\rho\left(T_k\right) , A_k\lceil \rho\left(T_k\right) \rfloor \right) - \beta\mathbb{D}_\mathrm{KL}\left(\pi_\theta||\pi_{\theta_\text{old}}\right) \big)\right],
\end{align}
where $A_k$ is the advantage score computed by standardizing the rewards within the group and $\lceil \cdot \rfloor$ denotes ratio clipping. This forces the model to ``self-correct'' by contrasting better geometric generations against worse ones for the same input.

\subsubsection{Hierarchical Reward Modulation}
Designing a dense reward signal is crucial for stable RL training. We propose a hierarchical reward function $R(T)$ that decomposes the objective into three levels of abstraction:

\begin{enumerate}
\item \textbf{Validity Check ($R_\mathrm{val}$):} A binary reward ensuring the output is valid JSON and able to form a watertight closed polygon.

\item \textbf{External Geometry ($R_\mathrm{ext}$):} The Intersection-over-Union between the predicted and ground truth external polygons. To prevent the model from optimizing internal details within a broken boundary, we use $R_{ext}$ to compute a gating factor $\alpha$:
\begin{equation}
    \alpha = \begin{cases} 
        0.1 & \text{if } R_\text{ext} < 0.3 \\
        \mathrm{lerp}(0.1, 1.0, \frac{R_\text{ext}-0.3}{0.7-0.3}),  & \text{if } 0.3 \le R_\text{ext} < 0.7 \\
        1.0 & \text{if } R_\text{ext} \ge 0.7 
    \end{cases},
\end{equation}
where $\mathrm{lerp}(a, b, t)$ represents the linear interpolation between $a$ and $b$ weighted by factor $t$.
\item \textbf{Internal Structure ($R_\mathrm{int}$):} This term evaluates F1 score and IoU metric of internal walls, openings, and room loops. Crucially, this reward is modulated by $\alpha$ (i.e., $\alpha \cdot R_\mathrm{int}$), strictly penalizing the model if the global boundary is incorrect.
\end{enumerate}

The final composite reward, guiding the model to prioritize global structural correctness before refining local internal details, is as follows:

\begin{equation}
\begin{split}
    R(T) = w_\mathrm{val}R_\mathrm{val} + w_\text{ext}R_\text{ext} + \alpha w_\mathrm{int}R_\mathrm{int},
\end{split}
\end{equation}
where the balancing coefficients are empirically set to $w_\mathrm{val}=0.1$, $w_\mathrm{ext}=0.5$, and $w_\mathrm{int}=0.4$.

\section{Experiments}
\label{sec:experiments}

\begin{figure}[t]
    \centering
    \subfloat[Geometry distribution \\of \textsc{FPBench-2K}.]{
        \includegraphics[width=0.48\linewidth]{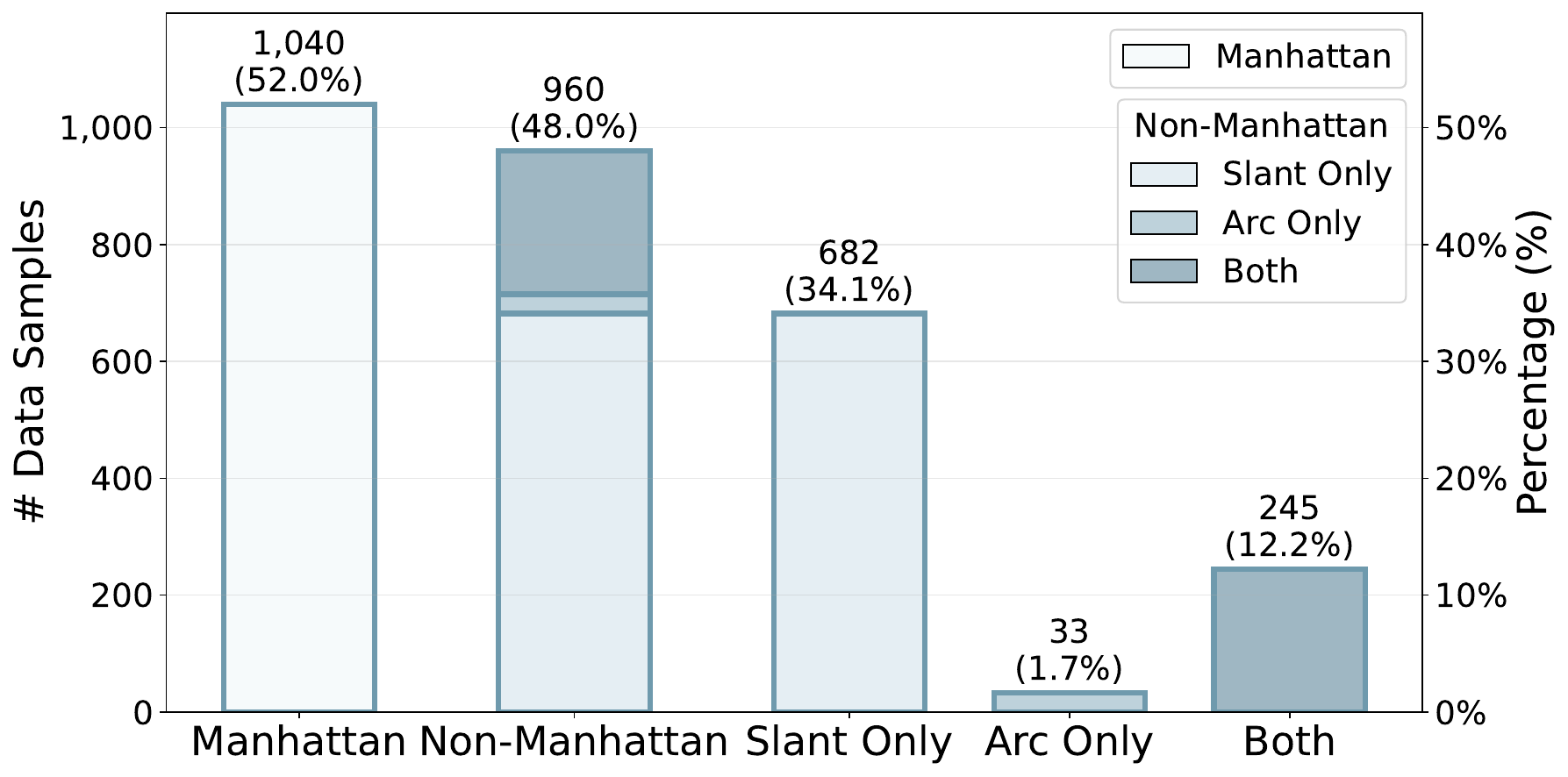}
        \label{fig:test_data_stats-a}
    }
    \subfloat[Primitive Count Distribution in \textsc{FPBench-2K}.]{
        \includegraphics[width=0.48\linewidth]{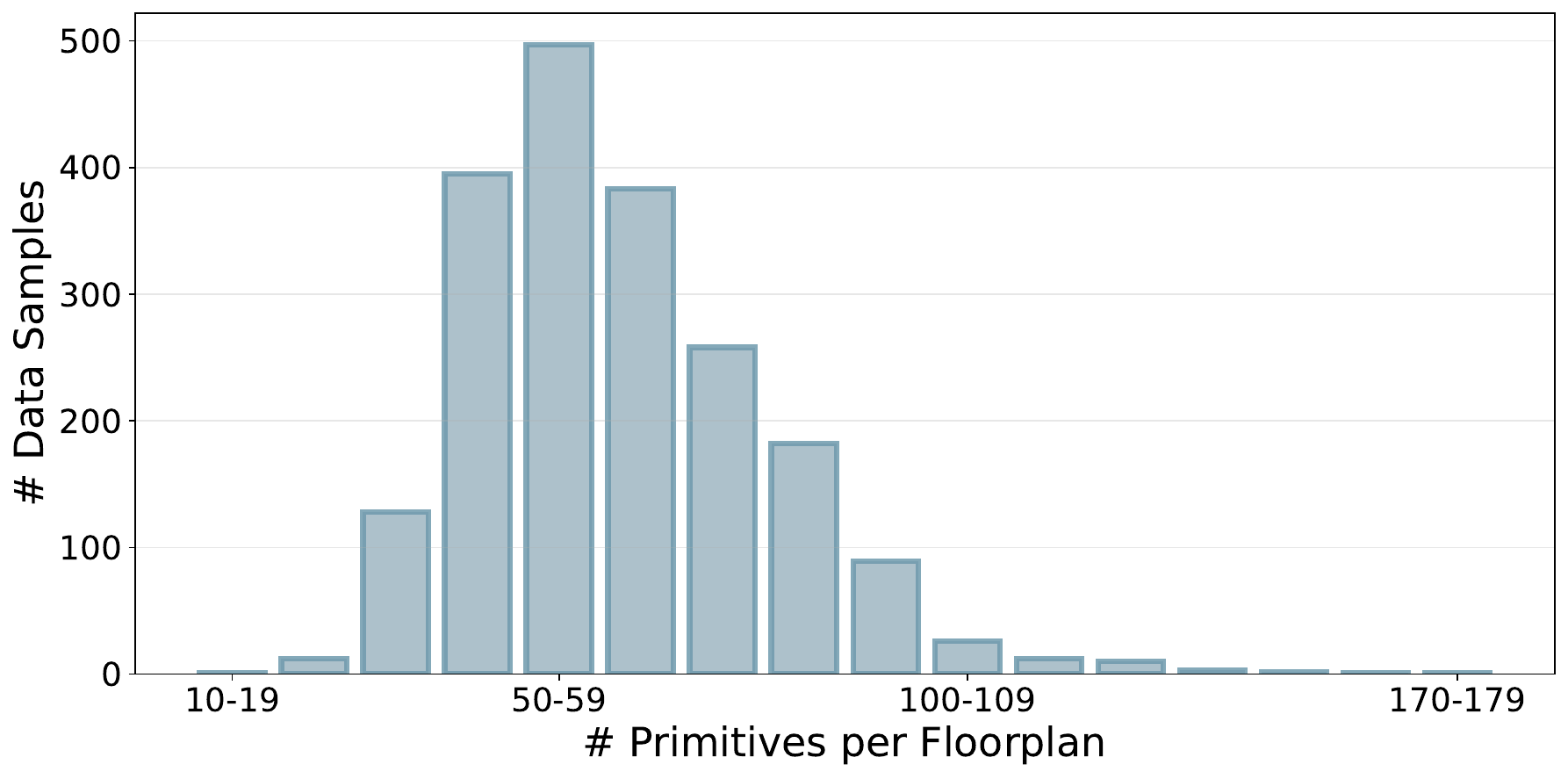}
        \label{fig:test_data_stats-b}
    }
    \caption{
        \textbf{Statistical Distribution of \textsc{FPBench-2K}.}
    }
    \label{fig:test_data_stats}
\end{figure}

Since existing benchmarks primarily focus on simplified rectilinear layouts, they fail to capture the full potential of our end-to-end vectorization paradigm. Therefore, we evaluate models on our custom benchmark to validate the effectiveness of the proposed architectural components, training stages, and data strategies.

\subsection{Experimental Setup}

\begin{table*}[t]
    \centering
    \caption{\textbf{Main Results on \textsc{FPBench-2K}.} We report structural validity ($\rho_\mathrm{val}$), geometric fidelity ($\mathrm{IoU}$) and semantic detection accuracy ($\mathrm{F1}$). The model demonstrates exceptional generalization on complex non-Manhattan geometries.}
    \label{tab:main_results}
    {
    \begin{tabular}{lcccccc}
        \toprule
        \textbf{Subset} & $\rho_\mathrm{val}$ (\%) & $\mathrm{IoU}_\mathrm{ext}$ & $\mathrm{IoU}_\mathrm{room}$ & $\mathrm{F1}_\mathrm{room}$ & $\mathrm{F1}_\mathrm{op}$ \\
        \midrule
        Manhattan & 97.02 & 0.9459 & 0.9089 & 0.8385 & 0.7739 \\
        Non-Manhattan & 95.10 & 0.9027 & 0.8738 & 0.8101 & 0.6894 \\
        \midrule
        \textbf{Overall} & \textbf{96.10} & \textbf{0.9252} & \textbf{0.8920} & \textbf{0.8249} & \textbf{0.7333} \\
        \bottomrule
    \end{tabular}
    }
\end{table*}

\paragraph{Dataset: \textsc{FPBench-2K}.}

To rigorously assess topological reasoning, we evaluate all models on \textsc{FPBench-2K}, a held-out set of 2,000 samples. We stratify this benchmark into two subsets:
\begin{itemize}
\item \textbf{Manhattan Subset (1040):} Layouts consisting strictly of orthogonal walls.
\item \textbf{Non-Manhattan Subset (960):} Complex layouts featuring slanted walls, curved balconies, and irregular polygons. This subset serves as the primary stress test for geometric generalization.
\end{itemize}

Furthermore, Figure~\ref{fig:test_data_stats}(b) confirms a wide range of primitive counts. Coupled with the increased non-Manhattan proportion, this provides a comprehensive stress test for geometric generalization across varying structural complexities.

\paragraph{Metrics.}
We employ a suite of vector-oriented metrics to measure structural validity:
\begin{itemize}
    \item  \textbf{Validity Rate ($\rho_\mathrm{val}$):} The percentage of generated samples that are both syntactically valid JSON and geometrically watertight (i.e., forming closed polygons without topological gaps).
    \item \textbf{Ext-IoU ($\mathrm{IoU}_\mathrm{ext}$) \& Room-IoU ($\mathrm{IoU}_\mathrm{room}$):} The Intersection-over-Union for the external boundary and individual rooms.
    \item \textbf{Room-F1 ($\mathrm{F1}_\mathrm{room}$) \& Opening-F1 ($\mathrm{F1}_\mathrm{op}$):} The harmonic mean of precision and recall for semantic element detection. A prediction is considered a true positive only if it satisfies both strict geometric alignment (e.g., $\text{IoU} > 0.5$) and correct semantic classification.
\end{itemize}

\paragraph{Implementation Details.}
We use Qwen2.5-VL-3B~\cite{bai2025qwen2} as the base model. Training is performed on 32$\times$H200 GPUs. Stage 1 uses the full \textsc{Floorplan-2M} dataset for 2 epochs, Stage 2 fine-tunes on \textsc{Floorplan-HQ-300K} for 10 epochs, and Stage 3 (GRPO) samples $G=8$ outputs per input with a KL coefficient of 0.01.

\begin{table*}[t]
    \centering
    \caption{\textbf{Ablation of Training Stages.} Stage 1 ensures structural generalization, Stage 2 refines visual alignment, and Stage 3 (GRPO) enforces strict geometric constraints.}
    \label{tab:stages}
    {
    \begin{tabular}{lcccccc}
        \toprule
        \textbf{Configuration} & $\rho_\mathrm{val}$ (\%) & $\mathrm{IoU}_\mathrm{ext}$ & $\mathrm{IoU}_\mathrm{room}$ & $\mathrm{F1}_\mathrm{room}$ & $\mathrm{F1}_\mathrm{op}$ \\
        \midrule
        (a) SFT Baseline (Stage 1 + 2) & 90.20 & 0.8567 & 0.8521 & 0.7945 & 0.7073 \\
        (b) \qquad \textit{w/o Quality Annealing (Stage 1 Only)} & 67.25 & 0.5238 & 0.4692 & 0.3873 & 0.3945 \\
        (c) \qquad \textit{w/o Structural Grounding (Stage 2 Only)} & 85.10 & 0.7598 & 0.7346 & 0.6674 & 0.6458 \\
        \midrule
        (d) \textbf{Ours w/ GRPO (Stage 1+2+3)} & \textbf{96.10} & \textbf{0.9252} & \textbf{0.8920} & \textbf{0.8249} & \textbf{0.7333} \\
        \bottomrule
    \end{tabular}
    }
\end{table*}

\subsection{Main Results: Geometric Generalization}

We report the performance of our final model after our three-stage training on the \textsc{FPBench-2K} test set. To assess geometric robustness, we break down performance by topological complexity (Manhattan vs. Non-Manhattan).

As shown in Table~\ref{tab:main_results}, \textsc{FloorplanVLM} achieves an exceptional 92.52\% external-boundary IoU ($\mathrm{IoU}_\mathrm{ext}$). Notably, even on the challenging non-Manhattan subset, the model maintains 90.27\% $\mathrm{IoU}_\mathrm{ext}$. The validity rate ($\rho_\mathrm{val}$) remains robust ($>$ 95\%) across both subsets. This confirms that our sequence modeling approach successfully learns the generalized grammar of architectural geometry, rather than merely overfitting to simple rectilinear patterns.

\subsection{Ablation Studies}

To validate our method, we conduct comprehensive ablation studies focusing on the training strategy, data scaling, and sequence representation.

\subsubsection{Impact of Progressive Training Strategy}
We analyze the contribution of each training stage. Table~\ref{tab:stages} compares the performance of models stopped at different phases.

\paragraph{Data Scale vs. Quality.} 
The ablation results in Table~\ref{tab:stages} offer a nuanced view of the training dynamics:
\begin{itemize}
    \item \textbf{Quality drives Topological Validity.} Comparing Table~\ref{tab:stages}(b) and (c), the model trained on the noisy 2M dataset (b) fails to learn loop closure ($\rho_\mathrm{val}=67.25\%$) because it overfits to the \textit{coordinate misalignment} and topological errors inherent in the raw screenshot data. In contrast, the pixel-aligned Stage 2 data (c) explicitly teaches the model the physical constraint of ``watertightness,'' boosting the validity to 85.10\%.
        
    \item \textbf{Scale drives Visual Generalization.} However, high-quality data alone reaches a performance ceiling. Comparing Table~\ref{tab:stages}(c) and (a), adding Stage 1 yields \textit{a substantial 0.10 gain} in $\mathrm{IoU}_\mathrm{ext}$ (0.76 $\to$ 0.86) and further improves validity to 90.20\%. This indicates that Stage 1 acts as \textit{structural grounding}: it exposes the model to a vast diversity of wall textures and layout structures, enabling it to ``recognize'' walls robustly in complex scenes. Stage 2 then refines this robust perception into topologically valid, pixel-precise vectors.
\end{itemize}

\begin{figure}[t]
    \centering
    \includegraphics[width=\linewidth]{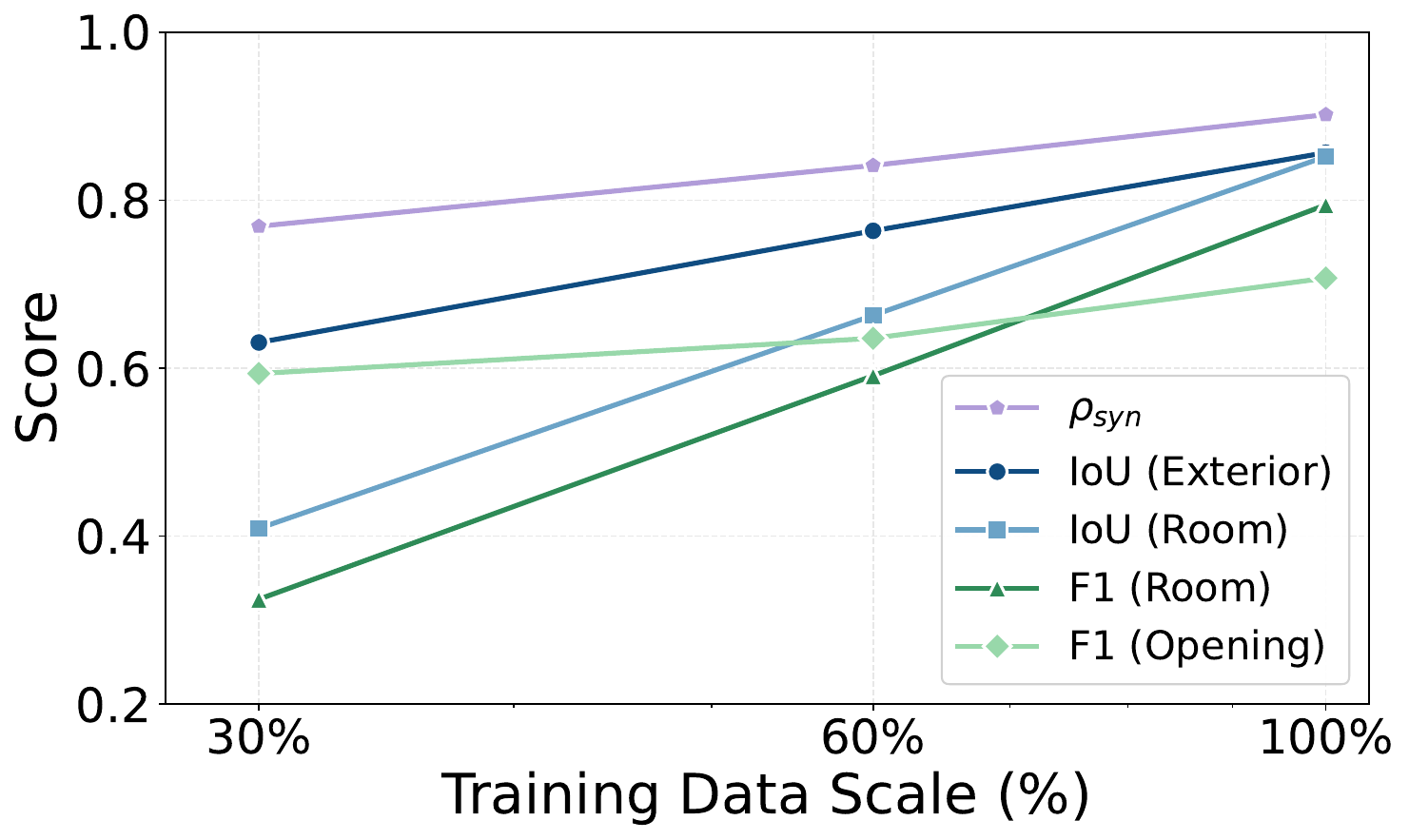}
    \caption{\textbf{Model Performance vs. Data Scale.} We evaluate the performance of Stage 1 \& 2 models trained on varying subsets (30\%, 60\%, and 100\%) of the \textsc{Floorplan-2M} and \textsc{Floorplan-HQ-300K} datasets. The x-axis is shown on a log scale.
    }
    \label{fig:data_scaling}
\end{figure}

\paragraph{The Leap with GRPO.} 
Finally, the comparison between Table~\ref{tab:stages}(a) and (d) highlights the limitations of SFT. While SFT achieves decent performance ($\rho_\mathrm{val}=90.20\%$), it operates via probabilistic imitation; optimizing token likelihood does not strictly penalize small geometric drifts that break topology.

Table~\ref{tab:stages}(d) shows that GRPO  overcomes this by introducing \textit{geometric enforcement}. By directly optimizing non-differentiable rewards ($R_\mathrm{val}$, $R_\mathrm{ext}$ and $R_\mathrm{int}$), GRPO forces the model to transition from simply ``mimicking'' valid floorplans to actively satisfying hard physical constraints. This results in a significant leap: a \textbf{+5.9\%} boost in validity and a \textbf{+0.07} surge in Ext-IoU, confirming that RL is the decisive factor for achieving engineering-grade precision.

\subsubsection{Data Scaling Laws}
Figure~\ref{fig:data_scaling} illustrates the trajectory of model performance as the training data volume increases from 30\% to 100\%. We observe a distinct log--linear improvement in geometric metrics, which yields two critical insights:

\begin{itemize}
\item \textbf{Unsaturated Capacity:} The performance curve shows no sign of plateauing even at 2M samples. This trend suggests that the pixels-to-sequence paradigm has sufficient capacity to absorb far more architectural patterns, and that current performance is likely limited by data scale rather than model architecture.
\item \textbf{Benefit of Structural Diversity:} Since our dataset is constructed via \textit{structure-aware clustering} rather than random sampling, increasing data volume effectively introduces more long-tail topological variants rather than mere redundancy. This confirms that scaling diverse data is a direct path to improving geometric generalization.
\end{itemize}

\subsubsection{Format Selection: Code vs. JSON}

We investigated the optimal sequence representation by comparing a Python-DSL against our JSON schema. To ensure a fair comparison, the Python variant also utilized keyword arguments (e.g., \texttt{Wall(width=10, ...)}) to maintain semantic explicitness.

\begin{table}[h]
    \centering
    \caption{\textbf{Format Comparison.} JSON outperforms Python despite having a longer sequence length. This can be attributed to the model's stronger pre-training alignment with JSON syntax.}
    \label{tab:format}
    \small
    \setlength{\tabcolsep}{1.8pt}
    \begin{tabular}{lcccccc}
        \toprule
        \textbf{Format}  &\textbf{Len.} $\downarrow$ & $\rho_\mathrm{val}$ (\%) & $\mathrm{IoU}_\mathrm{ext}$ & $\mathrm{IoU}_\mathrm{room}$ & $\mathrm{F1}_\mathrm{room}$ & $\mathrm{F1}_\mathrm{op}$ \\
        \midrule
        Python Code & \textbf{2038.4} & 88.15 & 0.8555 & 0.8050 & 0.8079 & 0.6407 \\
        \textbf{JSON (Ours)} & 3095.2 & \textbf{90.20} & \textbf{0.8567} & \textbf{0.8521} & \textbf{0.8140} & \textbf{0.7073}\\
        \bottomrule
    \end{tabular}
\end{table}

\paragraph{Pre-training Alignment outweighs Token Efficiency.}
As shown in Table~\ref{tab:format}, the Python format is significantly more compact ($\sim$2,000 tokens) than JSON ($\sim$3,000 tokens). Since both formats utilize keyword arguments, the performance gap cannot be attributed to semantic ambiguity.

Instead, we identify \textit{pre-training bias} as the decisive factor. Foundation models like Qwen2.5-VL are heavily pre-trained on web-scale data, where JSON is the \textit{de facto} standard for structured data interchange. Consequently, the model possesses a strong inductive bias toward JSON syntax constraints. In contrast, our specific Python DSL, while syntactically valid, represents a rare distribution in the pre-training corpus. This makes the model less stable when generating long Python sequences, leading to lower validity ($\rho_\mathrm{val}$) compared to the ``native-feeling'' JSON format.

\subsection{Qualitative Results}

\begin{figure*}[t]
    \centering
    \includegraphics[width=\textwidth]{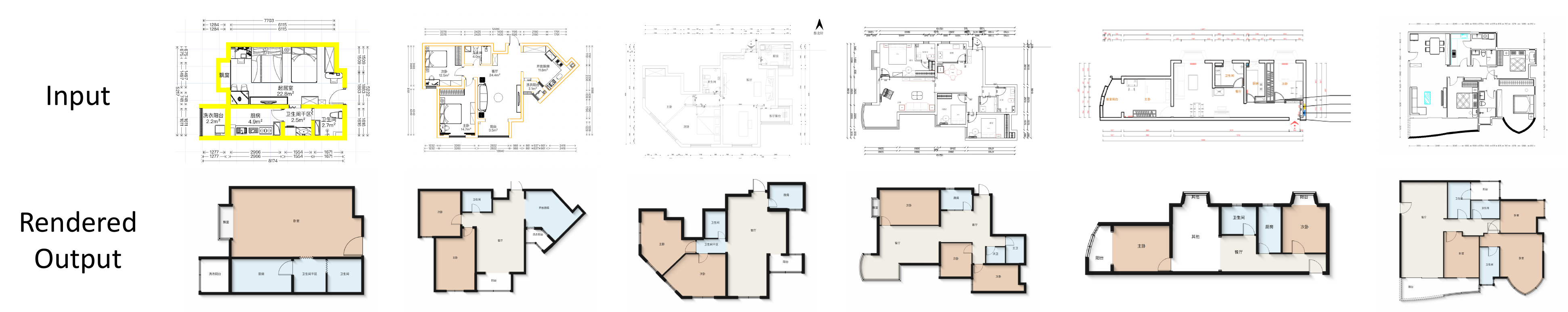}
    \caption{
        \textbf{Qualitative Visualization on \textsc{FPBench-2K}.} 
        We showcase the vectorization results of \textsc{FloorplanVLM} across diverse architectural styles. The model successfully reconstructs complex non-Manhattan geometries, including slanted walls (cols.~2--3) and curved balconies (cols.~4--6), while maintaining strict topological connectivity in the generated JSON. Note that the output is not a raster segmentation mask, but a fully parametric vector graph rendered for visualization. (Zoom in for best view.)
    }
    \label{show_cases}
\end{figure*}

\begin{figure}[t]
    \centering
    \includegraphics[width=0.48\textwidth]{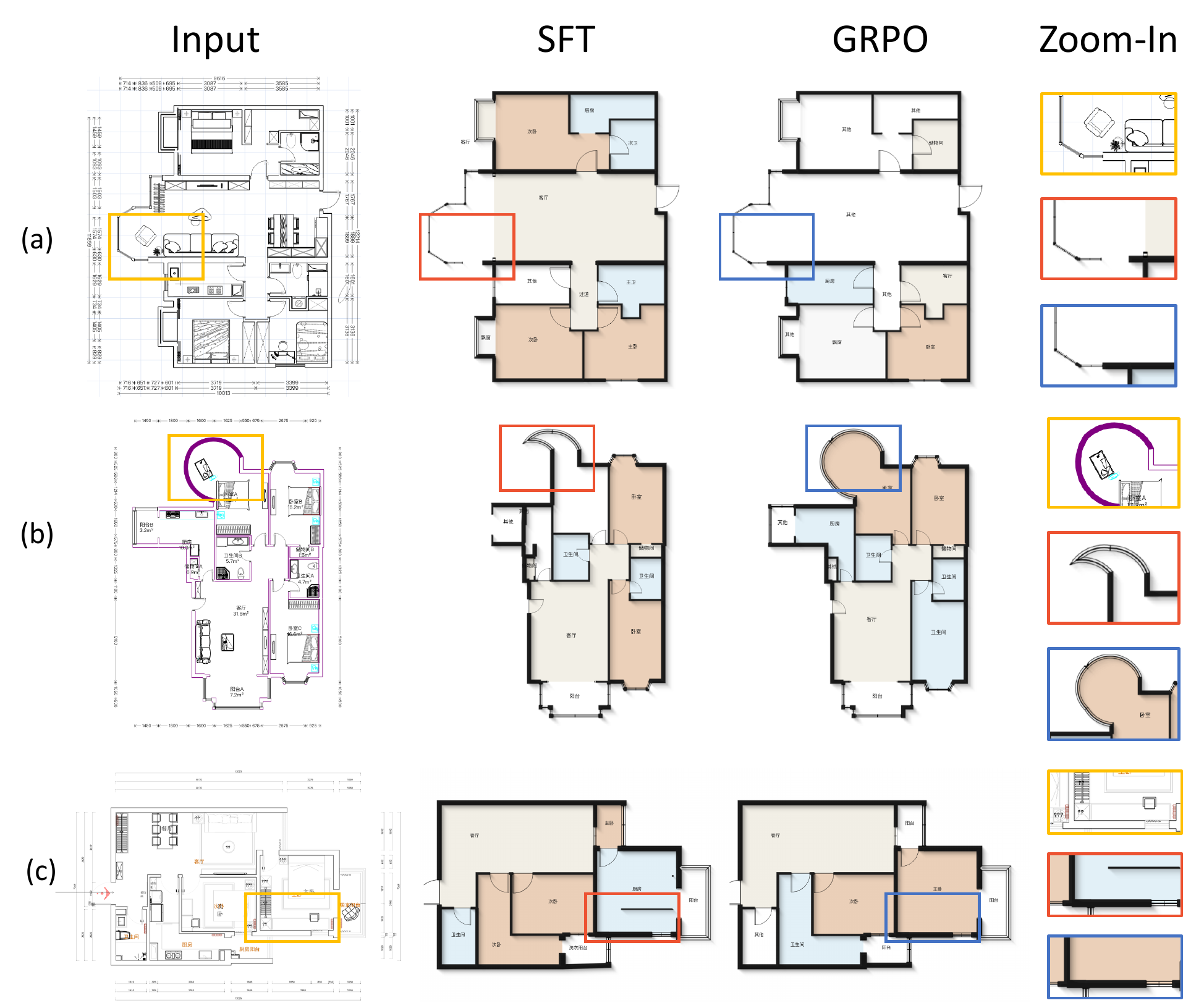}
    \caption{
        \textbf{Visual Comparison: SFT vs. GRPO.} 
        We analyze specific failure cases of the SFT baseline (Stage 2) and their correction by GRPO (Stage 3). 
        \textbf{(a)} Topological Gaps: SFT leaves unclosed gaps at wall junctions (orange box). 
        \textbf{(b)} Curvature Mismatch: SFT detects the curve but predicts an inaccurate curvature parameter $\kappa$, causing deviation from the visual boundary. 
        \textbf{(c)} Structural Hallucination: SFT incorrectly identifies or positions internal walls, missing the ground truth layout. 
        In all cases, GRPO strictly aligns geometric primitives with image evidence.
        (Best view in color and zoomed in.)
    }
    \label{fig:grpo_sft}
\end{figure}

We present qualitative examples to visually substantiate the quantitative metrics reported in Table~\ref{tab:main_results}.

\paragraph{Robustness to Noise and Complexity.} 
Figure~\ref{show_cases} demonstrates the model's capability to distill structural logic from noisy raster inputs. Despite the presence of interfering elements like furniture, dimension lines, and watermarks, \textsc{FloorplanVLM} accurately grounds visual patterns to geometric primitives. Notably, the model generalizes well to non-Manhattan topologies, precisely parsing the slanted walls (Col.~2) and curved boundaries (Col.~5) that typically confound traditional rule-based heuristics.

\paragraph{SFT vs. GRPO: Precision beyond Probability.}
Figure~\ref{fig:grpo_sft} highlights the critical role of Reinforcement Learning in achieving engineering-grade precision. While SFT provides a strong initialization, it operates on token likelihood, which does not necessarily correlate with pixel-perfect alignment. 

\begin{itemize}
    \item \textbf{Topological Closure (Row a):} SFT effectively captures the global shape but often fails at local corner connectivity, leaving ``hanging'' wall endpoints. This occurs because token-level likelihood objectives do not explicitly penalize small coordinate drifts that break topology. GRPO resolves this by integrating the binary validity reward ($R_\mathrm{val}$), effectively forcing the model to ``snap'' endpoints together to form watertight polygons.
    
    \item \textbf{Curvature Refinement (Row b):} In curved regions, SFT often correctly predicts the \textit{presence} of an arc but fails to estimate the exact curvature parameter $\kappa$. This results in a curve that ``looks'' roughly correct but deviates from the underlying pixel footprint. GRPO corrects this by directly optimizing the IoU reward, forcing the continuous parameter $\kappa$ to tightly fit the visual boundary.
    
    \item \textbf{Structural Integrity (Row c):} SFT struggles with ambiguous internal structures, leading to \textit{geometric hallucinations} where internal walls are either missed or incorrectly positioned. By incorporating the validity-gated reward ($R_\mathrm{int}$), GRPO effectively suppresses these false positives, ensuring that every generated wall is strictly grounded in image evidence.
\end{itemize}

In summary, while SFT learns the \textit{syntax} of floorplans, GRPO is essential for enforcing the \textit{physical constraints} and \textit{parametric accuracy} required for downstream architectural applications.

\section{Conclusion}

In this work, we present \textsc{FloorplanVLM}, a framework that reformulates floorplan vectorization from a traditional detection-assembly pipeline into an end-to-end image-conditioned sequence generation task. By leveraging the semantic reasoning power of large Vision-Language Models, our approach directly outputs structured, render-ready JSON, streamlining the workflow for architectural design.

Our ablation studies reveal two critical insights for engineering-grade generation: (1) scale drives perception: large-scale structural grounding on industrial data—despite inherent \textit{coordinate misalignment}—is indispensable for establishing a generalized visual foundation, enabling the model to recognize diverse topologies beyond simple layouts. (2) RL enforces geometric logic: while SFT mimics data distribution, it struggles with pixel-perfect precision. Integrating GRPO bridges this gap, transforming probabilistic token generation into a process that actively satisfies hard geometric constraints (e.g., watertight loop closure).

Evaluated on our newly established \textsc{FPBench-2K}, \textsc{FloorplanVLM} achieves strong performance, particularly on complex non-Manhattan geometries, demonstrating robust topological reasoning for vector graphics generation.

\section*{Limitations and Future Work}

While \textsc{FloorplanVLM} demonstrates strong performance, we acknowledge limitations that point towards future research directions:

\begin{itemize}
\item \textbf{Inference Latency:} The autoregressive generation of long JSON sequences (averaging $\sim$3,000 tokens) using a 3B-parameter model incurs higher latency than traditional lightweight CNNs. Future engineering efforts will focus on model quantization and speculative decoding to enable real-time deployment on edge devices.
    
\item \textbf{Discrete Precision Bottleneck:} Our strategy of quantizing coordinates to a fixed [0, 1024] integer grid balances training stability with accuracy, but it imposes a theoretical limit on precision compared to continuous floating-point regression. Exploring hybrid discrete-continuous heads could further enhance engineering-grade fidelity.
    
\item \textbf{Towards Interactive Agents:} Currently, the model operates as a one-shot translator. Moving forward, we aim to extend this framework to support multi-turn editing, enabling designers to iteratively refine layouts via natural language (e.g., ``Move this wall 0.5m to the right''), thereby unlocking the full potential of human-AI collaboration in architecture.
\end{itemize}

\section*{Ethical Statement}

We affirm that our research adheres to ethical guidelines regarding data privacy and usage. The \textsc{Floorplan-2M} dataset originates from an industrial interior design platform, where all personally identifiable information (PII), including homeowner names and specific geospatial coordinates, has been rigorously anonymized and desensitized prior to processing.

Regarding broader societal impact, while \textsc{FloorplanVLM} automates the labor-intensive task of vectorization, it is designed as an assistive tool to augment, rather than replace, human expertise. By automating tedious tracing work, we aim to free architects and designers to focus on high-level creative decision-making. We are also committed to releasing our benchmark (\textsc{FPBench-2K}) to promote transparency and reproducibility in the community.

\appendix

\section*{Acknowledgments}

We sincerely thank the architectural designers at Beike for their tremendous effort in manually annotating and verifying the high-fidelity floorplan dataset. We are also grateful to the Beike AI Infrastructure Team for providing the essential GPU resources. This research would not have been possible without the open-source community, particularly the Qwen team for the Qwen2.5-VL model.

\bibliographystyle{named}
\bibliography{ijcai26}

\end{document}